\documentclass[nonacm,sigconf]{acmart} 

\renewcommand\footnotetextcopyrightpermission[1]{} 
\usepackage{multirow}

\AtBeginDocument{%
  \providecommand\BibTeX{{%
    \normalfont B\kern-0.5em{\scshape i\kern-0.25em b}\kern-0.8em\TeX}}}

\begin{document}

\title[]{CareFall: Automatic Fall Detection through \\ Wearable Devices and AI Methods}


\author{Juan Carlos Ruiz-Garcia}
\affiliation{%
  \institution{Universidad Autonoma de Madrid}
  \city{Madrid}
  \country{Spain}}
\email{juanc.ruiz@uam.es}

\author{Ruben Tolosana}
\affiliation{%
  \institution{Universidad Autonoma de Madrid}
  \city{Madrid}
  \country{Spain}}
\email{ruben.tolosana@uam.es}

\author{Ruben Vera-Rodriguez}
\affiliation{%
  \institution{Universidad Autonoma de Madrid}
  \city{Madrid}
  \country{Spain}}
\email{ruben.vera@uam.es}

\author{Carlos Moro}
\affiliation{%
  \institution{Cartronic Group}
  \city{Madrid}
  \country{Spain}}
\email{cmoro@cartronic.es}

\renewcommand{\shortauthors}{}

\begin{abstract}
    The aging population has led to a growing number of falls in our society, affecting global public health worldwide. This paper presents CareFall, an automatic Fall Detection System (FDS) based on wearable devices and Artificial Intelligence (AI) methods. CareFall considers the accelerometer and gyroscope time signals extracted from a smartwatch. Two different approaches are used for feature extraction and classification: \textit{i)} threshold-based, and \textit{ii)} machine learning-based. Experimental results on two public databases show that the machine learning-based approach, which combines accelerometer and gyroscope information, outperforms the threshold-based approach in terms of accuracy, sensitivity, and specificity. This research contributes to the design of smart and user-friendly solutions to mitigate the negative consequences of falls among older people.
\end{abstract}




\begin{teaserfigure}
  \centering
  \includegraphics[width=16cm,height=4.5cm]{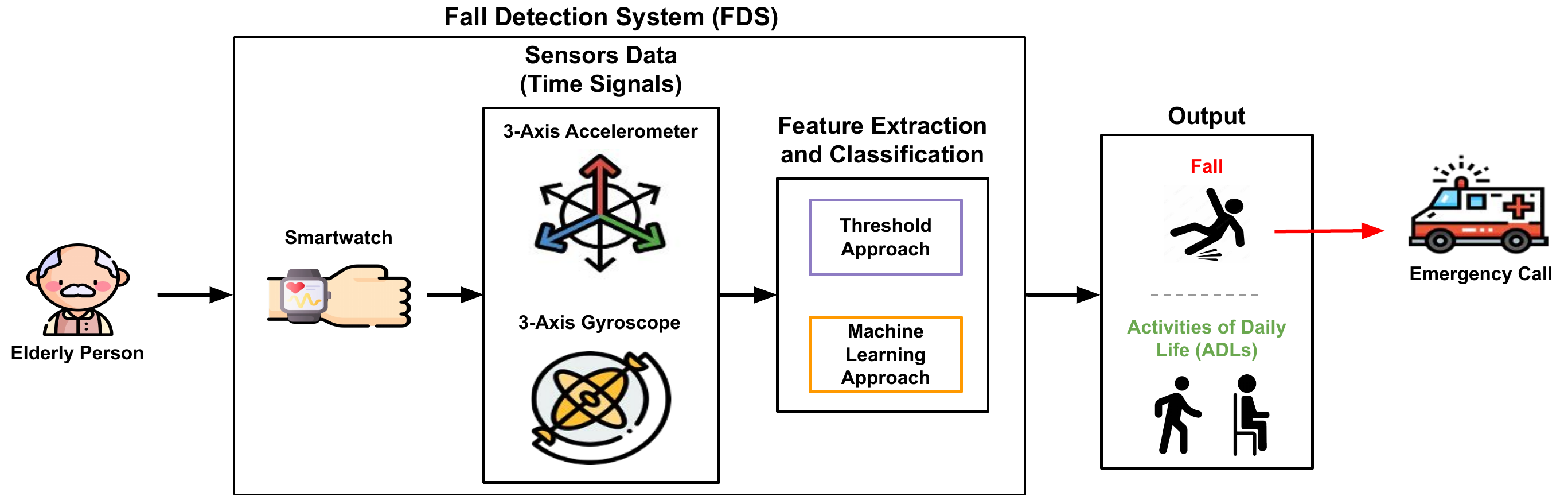}
  \caption{Representation of CareFall, an automatic Fall Detection System (FDS) based on wearable devices and AI methods.}
  \label{fig:figure1}
\end{teaserfigure}


\maketitle

\begin{table*}[!]
\caption{Selected public databases to train and evaluate CareFall.}
\label{tab:databases}
\resizebox{\linewidth}{!}{%
    \begin{tabular}{cccccccc}
    \textbf{\begin{tabular}[c]{@{}c@{}}Database\\ (Year) [Ref.]\end{tabular}}\rule{0pt}{20pt} & \textbf{\begin{tabular}[c]{@{}c@{}}\# Participants\\ (Females/Males)\end{tabular}} & \multicolumn{1}{c}{\textbf{\begin{tabular}[c]{@{}c@{}}Participant \\ Information\end{tabular}}} & \textbf{\begin{tabular}[c]{@{}c@{}}\# Tasks\\ (ADLs/Falls)\end{tabular}} & \textbf{\begin{tabular}[c]{@{}c@{}}\# Samples\\ (ADLS/Falls)\end{tabular}} & \textbf{\begin{tabular}[c]{@{}c@{}}\# Sensor\\ Position\end{tabular}} & \textbf{\begin{tabular}[c]{@{}c@{}}Captured\\ Signals\end{tabular}} & \textbf{\begin{tabular}[c]{@{}c@{}}\# Sampling\\ Rate (Hz)\end{tabular}} \\[+10pt] \hline
    \multicolumn{1}{c|}{\begin{tabular}[c]{@{}c@{}}Erciyes University\\ (2014)~\cite{Ozdemir2014}\end{tabular}} & \begin{tabular}[c]{@{}c@{}}17\\ (7/10)\end{tabular} & \begin{tabular}[c]{@{}l@{}}Age: 19-27 years\\ Height: 157-184 cm\\ Weight: 47-92 kg\end{tabular} & \begin{tabular}[c]{@{}c@{}}36\\ (16/20)\end{tabular} & \begin{tabular}[c]{@{}c@{}}3060\\ (1360/1700)\end{tabular} & \begin{tabular}[c]{@{}l@{}}Wrist (right)\end{tabular} & \begin{tabular}[c]{@{}c@{}}Acc: 3-axis\\ Gyr: 3-axis\end{tabular} & 25 \\ \hline
    \multicolumn{1}{c|}{\begin{tabular}[c]{@{}c@{}}UMAFall\\ (2016)~\cite{Casilari2017}\end{tabular}} & \begin{tabular}[c]{@{}c@{}}17\\ (7/10)\end{tabular} & \begin{tabular}[c]{@{}l@{}}Age: 18-55 years\\ Height: 155-195 cm\\ Weight: 50-93 kg\end{tabular} & \begin{tabular}[c]{@{}c@{}}11\\ (8/3)\end{tabular} & \begin{tabular}[c]{@{}c@{}}531\\ (322/209)\end{tabular} & \begin{tabular}[c]{@{}l@{}}Wrist (left)\end{tabular} & \begin{tabular}[c]{@{}c@{}}Acc: 3-axis\\ Gyr: 3-axis\end{tabular} & 20\\ \hline
    \end{tabular}
}
\end{table*}

\section{Introduction}



Population aging is increasing worldwide. The World Health Organization considers falls among the elderly to be a major global public health challenge~\cite{WHO2017}. In fact, falls can adversely affect the quality of life in older people, causing them serious physical, psychological, and social consequences, such as contusions, fractures, trauma, motor and neurological damage, or even death~\cite{Ambrose2013, Delahoz2014, Zhang2015}. For this reason, it is crucial the design and deployment of user-friendly technologies to detect falls. 


In recent years, solutions such as the Personal Emergency Response System (PERS) have been proposed~\cite{Bourke2007}. PERS is a manual system whereby a person, after falling to the ground, must press a warning button (usually in a pendant or bracelet), and an emergency team is immediately dispatched to provide assistance. However, this system might not be a good solution in some cases, e.g., if the person has fainted or lost consciousness due to the fall and can not press the emergency button.

To overcome the limitations of PERS, a wide variety of Fall Detection Systems (FDS) have been proposed in the last decade, providing automatic and user-friendly solutions for elderly people~\cite{Delahoz2014, Rashidi2013}. Most FDS are based on wearable devices~\cite{RomeroTapiador2023}, such as belts or bracelets with accelerometer sensors~\cite{Sucerquia2017, MartinezVillasenor2019, Kwolek2014}, image-based devices, such as indoor surveillance cameras~\cite{Galvao2021, Mastorakis2018}, or smartphones~\cite{Guvensan2017, Mellone2012, Kau2015}, among many others.


This paper presents CareFall, an automatic FDS based on wearable devices and Artificial Intelligence (AI) methods. Fig.~\ref{fig:figure1} provides a graphical representation of CareFall. CareFall considers a scenario where the  smartwatch is positioned on the wrist, acquiring information related to its inertial sensors, such as the 3-axis accelerometer and gyroscope~\cite{DelgadoSantos2022a, DelgadoSantos2022b}, or heart rate monitor~\cite{HernandezOrtega2020a, HernandezOrtega2020b}. Once the information is acquired by the smartwatch, the time signals (accelerometer and gyroscope signals) are used for feature extraction and classification. Two different approaches are considered: \textit{i)} threshold-based, and \textit{ii)} machine learning-based. In case the FDS detects a fall, it automatically warns the emergency services. 

\section{Methods}\label{sec:methods}


CareFall considers two of the most popular methods for fall detection in the literature~\cite{Schwickert2013}. They are fed to the 3-axis time signals of accelerometer and gyroscope sensors. The sampling frequency of the smartwatch is between 20-25Hz. For a simple and real-time analysis, we consider separate time windows of 1 minute.

\begin{enumerate}
    \item \textbf{Threshold-based}: it is one of the simplest and least computationally expensive solutions to detect a fall. It is based on the extraction of additional time signals from the original accelerometer and gyroscope ones such as the Signal Magnitude Vector (SMV), the Fall Index (FI), and the Absolute Vertical Direction (AVD), among others~\cite{Casilari2015, Casilari2017a}. After that, a specific threshold is defined for each time sequence. In case the instant value of the time sequence surpasses the threshold, the output of the system would be fall. It is important to highlight that, in case of considering several time signals in the analysis (e.g., SMV, FI, and AVD), the final output of the system would be based on the majority voting of all the time signals considered. 

    \item \textbf{Machine Learning-based}: this approach automatically learns the discriminative patterns for the task using data. From the original 6 time signals (3-axis accelerometer and gyroscope) and 2 additional time signals (SMV of the accelerometer and gyroscope), we extract the following 11 global features per time window (1 minute) related to statistical information: Mean, Variance, Median, Delta, Standard Deviation, Maximum Value, Minimum Value, 25th Percentile, 75th Percentile, Power Spectral Density (PSD), and Power Spectral Entropy (PSE). In total, we obtain a feature vector with 44 global features related to the accelerometer information and 44 global features related to the gyroscope. 
    
    Once we have the feature vector with the 88 global features, we train machine learning classifiers for the task of fall detection. The most widely used algorithms are K-Nearest Neighbor (KNN)~\cite{Ozdemir2014}, Support Vector Machine (SVM)~\cite{Dinh2009}, Gradient Boosting (GB)~\cite{Zurbuchen2021}, Random Forest (RF)~\cite{Zurbuchen2021}, and Artificial Neural Network (ANN)~\cite{Abbate2012}, among others. 
\end{enumerate}

\section{Experimental Setup}\label{sec:databases}


Two popular public databases are considered in the experimental framework of the paper: Erciyes Univesity~\cite{Ozdemir2014} and UMAFall~\cite{Casilari2017}. Table~\ref{tab:databases} shows the most relevant information from these databases: \textit{i)} the number of Activities of Daily Life (ADLs) such as walking, sitting, lying down, etc., and simulated falls (forward, backward, sideways, etc.); \textit{ii)} participant information (number, gender, height, weight, and age range); \textit{iii)} type of time signals captured (accelerometer and gyroscope); \textit{iv)} sensor position; and \textit{v)} the sampling rate. The main criteria for selecting these databases were the position of the sensor (wrist), the sampling rate of the sensors (20-25Hz), and the variability in the type of activities and falls.

Regarding the experimental protocol, both databases are divided into development (80\% of participants) and final evaluation (20\% remaining participants) datasets. As a result, different subjects are considered for the training and final evaluation of CareFall. Regarding metrics, we consider three popular metrics in the literature: Sensitivity (SE), Specificity (SP), and Accuracy. SE refers to the probability of detecting a fall, SP to the probability of detecting a non-fall (i.e., ADLs), and accuracy to the overall system performance.

\begin{table}[t]
    \caption{Results obtained in terms of accuracy, sensitivity, and specificity for Erciyes University and UMAFall databases over the final evaluation set. Results from both Threshold- and Machine Learning-based approaches are included.}
    \label{tab:results}
    \resizebox{0.5\textwidth}{!}{ 
    \begin{tabular}{cc|ccc}
     \multicolumn{1}{c|}{\multirow{3}{*}{\textbf{\begin{tabular}[c]{@{}c@{}}Erciyes University\\ (2014)~\cite{Ozdemir2014}\end{tabular}}}} & \multirow{2}{*}{\textbf{\begin{tabular}[c]{@{}c@{}}Threshold \\ Approach\\ (SMV)\end{tabular}}} & \multicolumn{3}{c}{\textbf{\begin{tabular}[c]{@{}c@{}}Machine Learning Approach\\ (RF)\end{tabular}}} \\ \cline{3-5} 
     \multicolumn{1}{c|}{} &  & \multicolumn{1}{c|}{\textbf{\begin{tabular}[c]{@{}c@{}}Accelerometer\\ (44 Features)\end{tabular}}} & \multicolumn{1}{c|}{\textbf{\textbf{\begin{tabular}[c]{@{}c@{}}Gyroscope\\ (44 Features)\end{tabular}}}} & \textbf{\begin{tabular}[c]{@{}c@{}}Acc + Gyr\\ (88 Features)\end{tabular}} \\ \hline
    \multicolumn{1}{c|}{\textbf{Accuracy}} & 77.3\% & \multicolumn{1}{c|}{97.2\%} & \multicolumn{1}{c|}{95.3\%} & 98.4\% \\ \hline
    \multicolumn{1}{c|}{\textbf{Sensitivity (SE)}} & 100.0\% & \multicolumn{1}{c|}{98.5\%} & \multicolumn{1}{c|}{97.9\%} & 98.9\% \\ \hline
    \multicolumn{1}{c|}{\textbf{Specificity (SP)}} & 68.4\% & \multicolumn{1}{c|}{95.8\%} & \multicolumn{1}{c|}{91.9\%} & 96.7\% \\ \hline \\

    \multicolumn{1}{c|}{\multirow{3}{*}{\textbf{\begin{tabular}[c]{@{}c@{}}UMAFall\\ (2016)~\cite{Casilari2017}\end{tabular}}}} & \multirow{2}{*}{\textbf{\begin{tabular}[c]{@{}c@{}}Threshold \\ Approach\\ (FI)\end{tabular}}} & \multicolumn{3}{c}{\textbf{\begin{tabular}[c]{@{}c@{}}Machine Learning Approach\\ (SVM)\end{tabular}}} \\ \cline{3-5} 
     \multicolumn{1}{c|}{} &  & \multicolumn{1}{c|}{\textbf{\begin{tabular}[c]{@{}c@{}}Accelerometer\\ (44 Features)\end{tabular}}} & \multicolumn{1}{c|}{\textbf{\textbf{\begin{tabular}[c]{@{}c@{}}Gyroscope\\ (44 Features)\end{tabular}}}} & \textbf{\begin{tabular}[c]{@{}c@{}}Acc + Gyr\\ (88 Features)\end{tabular}} \\ \hline
    \multicolumn{1}{c|}{\textbf{Accuracy}} & 75.8\% & \multicolumn{1}{c|}{93.9\%} & \multicolumn{1}{c|}{93.2\%} & 95.5\% \\ \hline
    \multicolumn{1}{c|}{\textbf{Sensitivity (SE)}} & 100.0\% & \multicolumn{1}{c|}{100.0\%} & \multicolumn{1}{c|}{97.3\%} & 98.6\% \\ \hline
    \multicolumn{1}{c|}{\textbf{Specificity (SP)}} & 66.3\% & \multicolumn{1}{c|}{91.6\%} & \multicolumn{1}{c|}{91.6\%} & 93.7\% \\ \hline
    \end{tabular}
    }
\end{table}

\section{Experimental Results}\label{sec:results}

Table~\ref{tab:results} (top) shows the results for the Erciyes University database over the final evaluation set. The results presented correspond to the best configuration of each fall detection approach. The results obtained in general (accuracy) with the threshold-based approach are significantly worse compared with the machine learning approach (77.3\% vs. 98.4\%), resulting in a higher number of false positives (no falls detected as falls). This trend can be observed by looking at the specificity (68.4\% vs. 96.7\%). Nevertheless, it is interesting to remark that the Threshold approach outperforms the Machine Learning approach in terms of sensitivity (100\% vs. 98.9\%), showing to be a simple but efficient approach to detecting falls. In addition, analysing the Machine Learning approach, we can see how the combination of accelerometer information (44 global features) and gyroscope information (44 global features) achieves the best results. 

Finally, we can also see in Table~\ref{tab:results} (bottom) the results achieved for the public UMAFall database. Similar conclusions are obtained, although better results are achieved on Erciyes database. This can be produced due to the quality of the device and the acquisition process. This seems to indicate that combining accelerometer and gyroscope information is a good practice for the fall detection task.


\section{Acknowledgments}

This work has been supported by projects: INTER-ACTION (PID2021-126521OBI00 MICINN/FEDER), HumanCAIC (TED2021-131787B-I00 MICINN), and Cartronic Group.

\bibliographystyle{ACM-Reference-Format}
\bibliography{refs}


\begin{thebibliography}{27}


\ifx \showCODEN    \undefined \def \showCODEN     #1{\unskip}     \fi
\ifx \showDOI      \undefined \def \showDOI       #1{#1}\fi
\ifx \showISBNx    \undefined \def \showISBNx     #1{\unskip}     \fi
\ifx \showISBNxiii \undefined \def \showISBNxiii  #1{\unskip}     \fi
\ifx \showISSN     \undefined \def \showISSN      #1{\unskip}     \fi
\ifx \showLCCN     \undefined \def \showLCCN      #1{\unskip}     \fi
\ifx \shownote     \undefined \def \shownote      #1{#1}          \fi
\ifx \showarticletitle \undefined \def \showarticletitle #1{#1}   \fi
\ifx \showURL      \undefined \def \showURL       {\relax}        \fi
\providecommand\bibfield[2]{#2}
\providecommand\bibinfo[2]{#2}
\providecommand\natexlab[1]{#1}
\providecommand\showeprint[2][]{arXiv:#2}

\bibitem[Abbate et~al\mbox{.}(2012)]%
        {Abbate2012}
\bibfield{author}{\bibinfo{person}{Stefano Abbate}, \bibinfo{person}{Marco
  Avvenuti}, \bibinfo{person}{Francesco Bonatesta}, \bibinfo{person}{Guglielmo
  Cola}, \bibinfo{person}{Paolo Corsini}, {and} \bibinfo{person}{Alessio
  Vecchio}.} \bibinfo{year}{2012}\natexlab{}.
\newblock \showarticletitle{{A Smartphone-Based Fall Detection System}}.
\newblock \bibinfo{journal}{\emph{Pervasive and Mobile Computing}}
  \bibinfo{volume}{8}, \bibinfo{number}{6} (\bibinfo{date}{Dec.}
  \bibinfo{year}{2012}), \bibinfo{pages}{883--899}.
\newblock
\urldef\tempurl%
\url{https://doi.org/10.1016/j.pmcj.2012.08.003}
\showDOI{\tempurl}


\bibitem[Ambrose et~al\mbox{.}(2013)]%
        {Ambrose2013}
\bibfield{author}{\bibinfo{person}{Anne~Felicia Ambrose}, \bibinfo{person}{Geet
  Paul}, {and} \bibinfo{person}{Jeffrey~M. Hausdorff}.}
  \bibinfo{year}{2013}\natexlab{}.
\newblock \showarticletitle{{{Risk Factors for Falls Among Older Adults: A
  Review of the Literature}}}.
\newblock \bibinfo{journal}{\emph{Maturitas}} \bibinfo{volume}{75},
  \bibinfo{number}{1} (\bibinfo{date}{March} \bibinfo{year}{2013}),
  \bibinfo{pages}{51--61}.
\newblock
\urldef\tempurl%
\url{https://doi.org/10.1016/j.maturitas.2013.02.009}
\showDOI{\tempurl}


\bibitem[Bourke et~al\mbox{.}(2007)]%
        {Bourke2007}
\bibfield{author}{\bibinfo{person}{A.K. Bourke}, \bibinfo{person}{J.V.
  O’Brien}, {and} \bibinfo{person}{G.M. Lyons}.}
  \bibinfo{year}{2007}\natexlab{}.
\newblock \showarticletitle{{Evaluation of a Threshold-Based Tri-Axial
  Accelerometer Fall Detection Algorithm}}.
\newblock \bibinfo{journal}{\emph{Gait \& Posture}} \bibinfo{volume}{26},
  \bibinfo{number}{2} (\bibinfo{date}{July} \bibinfo{year}{2007}),
  \bibinfo{pages}{194--199}.
\newblock
\urldef\tempurl%
\url{https://doi.org/10.1016/j.gaitpost.2006.09.012}
\showDOI{\tempurl}


\bibitem[Casilari and Oviedo-Jiménez(2015)]%
        {Casilari2015}
\bibfield{author}{\bibinfo{person}{Eduardo Casilari} {and}
  \bibinfo{person}{Miguel~A. Oviedo-Jiménez}.}
  \bibinfo{year}{2015}\natexlab{}.
\newblock \showarticletitle{{Automatic Fall Detection System Based on the
  Combined Use of a Smartphone and a Smartwatch}}.
\newblock \bibinfo{journal}{\emph{PLOS ONE}} \bibinfo{volume}{10},
  \bibinfo{number}{11} (\bibinfo{date}{Nov.} \bibinfo{year}{2015}),
  \bibinfo{pages}{1--11}.
\newblock
\urldef\tempurl%
\url{https://doi.org/10.1371/journal.pone.0140929}
\showDOI{\tempurl}


\bibitem[Casilari et~al\mbox{.}(2017a)]%
        {Casilari2017a}
\bibfield{author}{\bibinfo{person}{Eduardo Casilari},
  \bibinfo{person}{José-Antonio Santoyo-Ramón}, {and}
  \bibinfo{person}{José-Manuel Cano-García}.}
  \bibinfo{year}{2017}\natexlab{a}.
\newblock \showarticletitle{{Analysis of Public Datasets for Wearable Fall
  Detection Systems}}.
\newblock \bibinfo{journal}{\emph{Sensors}} \bibinfo{volume}{17},
  \bibinfo{number}{7} (\bibinfo{date}{June} \bibinfo{year}{2017}).
\newblock
\urldef\tempurl%
\url{https://doi.org/10.3390/s17071513}
\showDOI{\tempurl}


\bibitem[Casilari et~al\mbox{.}(2017b)]%
        {Casilari2017}
\bibfield{author}{\bibinfo{person}{Eduardo Casilari}, \bibinfo{person}{Jose~A.
  Santoyo-Ramón}, {and} \bibinfo{person}{Jose~M. Cano-García}.}
  \bibinfo{year}{2017}\natexlab{b}.
\newblock \showarticletitle{{UMAFall: A Multisensor Dataset for the Research on
  Automatic Fall Detection}}.
\newblock \bibinfo{journal}{\emph{Procedia Computer Science}}
  \bibinfo{volume}{110} (\bibinfo{date}{July} \bibinfo{year}{2017}),
  \bibinfo{pages}{32--39}.
\newblock
\urldef\tempurl%
\url{https://doi.org/10.1016/j.procs.2017.06.110}
\showDOI{\tempurl}


\bibitem[Delahoz and Labrador(2014)]%
        {Delahoz2014}
\bibfield{author}{\bibinfo{person}{Yueng~Santiago Delahoz} {and}
  \bibinfo{person}{Miguel~Angel Labrador}.} \bibinfo{year}{2014}\natexlab{}.
\newblock \showarticletitle{{{Survey on Fall Detection and Fall Prevention
  Using Wearable and External Sensors}}}.
\newblock \bibinfo{journal}{\emph{Sensors}} \bibinfo{volume}{14},
  \bibinfo{number}{10} (\bibinfo{date}{Oct.} \bibinfo{year}{2014}),
  \bibinfo{pages}{19806--19842}.
\newblock
\urldef\tempurl%
\url{https://doi.org/10.3390/s141019806}
\showDOI{\tempurl}


\bibitem[Delgado-Santos et~al\mbox{.}(2023)]%
        {DelgadoSantos2022a}
\bibfield{author}{\bibinfo{person}{Paula Delgado-Santos},
  \bibinfo{person}{Ruben Tolosana}, \bibinfo{person}{Richard Guest},
  \bibinfo{person}{Farzin Deravi}, {and} \bibinfo{person}{Ruben
  Vera-Rodriguez}.} \bibinfo{year}{2023}\natexlab{}.
\newblock \showarticletitle{{Exploring Transformers for Behavioural Biometrics:
  A Case Study in Gait Recognition}}.
\newblock \bibinfo{journal}{\emph{Pattern Recognition}} (\bibinfo{year}{2023}).
\newblock
\urldef\tempurl%
\url{https://doi.org/10.48550/arXiv.2206.01441}
\showDOI{\tempurl}


\bibitem[Delgado-Santos et~al\mbox{.}(2022)]%
        {DelgadoSantos2022b}
\bibfield{author}{\bibinfo{person}{Paula Delgado-Santos},
  \bibinfo{person}{Ruben Tolosana}, \bibinfo{person}{Richard Guest},
  \bibinfo{person}{Ruben Vera-Rodriguez}, \bibinfo{person}{Farzin Deravi},
  {and} \bibinfo{person}{Aythami Morales}.} \bibinfo{year}{2022}\natexlab{}.
\newblock \showarticletitle{{GaitPrivacyON: Privacy-Preserving Mobile Gait
  Biometrics using Unsupervised Learning}}.
\newblock \bibinfo{journal}{\emph{Pattern Recognition Letters}}
  \bibinfo{volume}{161} (\bibinfo{date}{Sept.} \bibinfo{year}{2022}),
  \bibinfo{pages}{30--37}.
\newblock
\urldef\tempurl%
\url{https://doi.org/10.1016/j.patrec.2022.07.015}
\showDOI{\tempurl}


\bibitem[Dinh et~al\mbox{.}(2009)]%
        {Dinh2009}
\bibfield{author}{\bibinfo{person}{Anh Dinh}, \bibinfo{person}{Daniel Teng},
  \bibinfo{person}{Li Chen}, \bibinfo{person}{Yang Shi}, \bibinfo{person}{Carl
  McCrosky}, \bibinfo{person}{Jenny Basran}, {and} \bibinfo{person}{Vanina~Del
  Bello-Hass}.} \bibinfo{year}{2009}\natexlab{}.
\newblock \showarticletitle{{Implementation of a Physical Activity Monitoring
  System for the Elderly People with Built-in Vital Sign and Fall Detection}}.
  In \bibinfo{booktitle}{\emph{Proceeding of the 6th International Conference
  on Information Technology: New Generations}}. \bibinfo{address}{Las Vegas,
  NV, USA}, \bibinfo{pages}{1226--1231}.
\newblock


\bibitem[Galvão et~al\mbox{.}(2021)]%
        {Galvao2021}
\bibfield{author}{\bibinfo{person}{Yves~M. Galvão}, \bibinfo{person}{Janderson
  Ferreira}, \bibinfo{person}{Vinícius~A. Albuquerque}, \bibinfo{person}{Pablo
  Barros}, {and} \bibinfo{person}{Bruno~J.T. Fernandes}.}
  \bibinfo{year}{2021}\natexlab{}.
\newblock \showarticletitle{{A Multimodal Approach Using Deep Learning for Fall
  Detection}}.
\newblock \bibinfo{journal}{\emph{{Expert Systems with Applications}}}
  \bibinfo{volume}{168} (\bibinfo{date}{April} \bibinfo{year}{2021}),
  \bibinfo{pages}{114226}.
\newblock
\urldef\tempurl%
\url{https://doi.org/10.1016/j.eswa.2020.114226}
\showDOI{\tempurl}


\bibitem[Guvensan et~al\mbox{.}(2017)]%
        {Guvensan2017}
\bibfield{author}{\bibinfo{person}{M.~Amac Guvensan}, \bibinfo{person}{A.~Oguz
  Kansiz}, \bibinfo{person}{N.~Cihan Camgoz}, \bibinfo{person}{H.~Irem
  Turkmen}, \bibinfo{person}{A.~Gokhan Yavuz}, {and} \bibinfo{person}{M.~Elif
  Karsligil}.} \bibinfo{year}{2017}\natexlab{}.
\newblock \showarticletitle{{An Energy-Efficient Multi-Tier Architecture for
  Fall Detection on Smartphones}}.
\newblock \bibinfo{journal}{\emph{{Sensors}}} \bibinfo{volume}{17},
  \bibinfo{number}{7} (\bibinfo{date}{June} \bibinfo{year}{2017}),
  \bibinfo{pages}{1--21}.
\newblock
\urldef\tempurl%
\url{https://doi.org/10.3390%2Fs17071487}
\showDOI{\tempurl}


\bibitem[Hernandez-Ortega et~al\mbox{.}(2020)]%
        {HernandezOrtega2020a}
\bibfield{author}{\bibinfo{person}{Javier Hernandez-Ortega},
  \bibinfo{person}{Roberto Daza}, \bibinfo{person}{Aythami Morales},
  \bibinfo{person}{Julian Fierrez}, {and} \bibinfo{person}{Ruben Tolosana}.}
  \bibinfo{year}{2020}\natexlab{}.
\newblock \showarticletitle{{Heart Rate Estimation from Face Videos for Student
  Assessment: Experiments on edBB}}. In \bibinfo{booktitle}{\emph{Proceedings
  of the IEEE 44th Annual Computers, Software, and Applications Conference
  (COMPSAC)}}. \bibinfo{publisher}{IEEE Computer Society},
  \bibinfo{address}{Los Alamitos, CA, USA}, \bibinfo{pages}{172--177}.
\newblock


\bibitem[Hernandez-Ortega et~al\mbox{.}(2021)]%
        {HernandezOrtega2020b}
\bibfield{author}{\bibinfo{person}{Javier Hernandez-Ortega},
  \bibinfo{person}{Ruben Tolosana}, \bibinfo{person}{Julian Fierrez}, {and}
  \bibinfo{person}{Aythami Morales}.} \bibinfo{year}{2021}\natexlab{}.
\newblock \showarticletitle{{DeepFakesON-Phys: DeepFakes Detection based on
  Heart Rate Estimation}}. In \bibinfo{booktitle}{\emph{Proceedings of the 35th
  AAAI Conference on Artificial Intelligence Workshops (AAAIw)}}.
  \bibinfo{address}{Vancouver, Canada}.
\newblock


\bibitem[Kau and Chen(2015)]%
        {Kau2015}
\bibfield{author}{\bibinfo{person}{Lih-Jen Kau} {and}
  \bibinfo{person}{Chih-Sheng Chen}.} \bibinfo{year}{2015}\natexlab{}.
\newblock \showarticletitle{{A Smart Phone-Based Pocket Fall Accident
  Detection, Positioning, and Rescue System}}.
\newblock \bibinfo{journal}{\emph{IEEE Journal of Biomedical and Health
  Informatics}} \bibinfo{volume}{19}, \bibinfo{number}{1} (\bibinfo{date}{June}
  \bibinfo{year}{2015}), \bibinfo{pages}{44--56}.
\newblock
\urldef\tempurl%
\url{https://doi.org/10.1109/JBHI.2014.2328593}
\showDOI{\tempurl}


\bibitem[Kwolek and Kepski(2014)]%
        {Kwolek2014}
\bibfield{author}{\bibinfo{person}{Bogdan Kwolek} {and} \bibinfo{person}{Michal
  Kepski}.} \bibinfo{year}{2014}\natexlab{}.
\newblock \showarticletitle{{Human Fall Detection on Embedded Platform Using
  Depth Maps and Wireless Accelerometer}}.
\newblock \bibinfo{journal}{\emph{Computer Methods and Programs in
  Biomedicine}} \bibinfo{volume}{117}, \bibinfo{number}{3}
  (\bibinfo{date}{Dec.} \bibinfo{year}{2014}), \bibinfo{pages}{489--501}.
\newblock
\urldef\tempurl%
\url{https://doi.org/10.1016/j.cmpb.2014.09.005}
\showDOI{\tempurl}


\bibitem[Martínez-Villaseñor et~al\mbox{.}(2019)]%
        {MartinezVillasenor2019}
\bibfield{author}{\bibinfo{person}{Lourdes Martínez-Villaseñor},
  \bibinfo{person}{Hiram Ponce}, \bibinfo{person}{Jorge Brieva},
  \bibinfo{person}{Ernesto Moya-Albor}, \bibinfo{person}{José
  Núñez-Martínez}, {and} \bibinfo{person}{Carlos Peñafort-Asturiano}.}
  \bibinfo{year}{2019}\natexlab{}.
\newblock \showarticletitle{{UP-Fall Detection Dataset: A Multimodal
  Approach}}.
\newblock \bibinfo{journal}{\emph{Sensors}} \bibinfo{volume}{19},
  \bibinfo{number}{9} (\bibinfo{date}{April} \bibinfo{year}{2019}),
  \bibinfo{pages}{1--28}.
\newblock
\urldef\tempurl%
\url{https://doi.org/10.3390%2Fs19091988}
\showDOI{\tempurl}


\bibitem[Mastorakis et~al\mbox{.}(2018)]%
        {Mastorakis2018}
\bibfield{author}{\bibinfo{person}{Georgios Mastorakis}, \bibinfo{person}{Tim
  Ellis}, {and} \bibinfo{person}{Dimitrios Makris}.}
  \bibinfo{year}{2018}\natexlab{}.
\newblock \showarticletitle{{Fall Detection Without People: A Simulation
  Approach Tackling Video Data Scarcity}}.
\newblock \bibinfo{journal}{\emph{{Expert Systems with Applications}}}
  \bibinfo{volume}{112} (\bibinfo{date}{Dec.} \bibinfo{year}{2018}),
  \bibinfo{pages}{125--137}.
\newblock
\urldef\tempurl%
\url{https://doi.org/10.1016/j.eswa.2018.06.019}
\showDOI{\tempurl}


\bibitem[Mellone et~al\mbox{.}(2012)]%
        {Mellone2012}
\bibfield{author}{\bibinfo{person}{S. Mellone}, \bibinfo{person}{C. Tacconi},
  \bibinfo{person}{L. Schwickert}, \bibinfo{person}{J. Klenk},
  \bibinfo{person}{C. Becker}, {and} \bibinfo{person}{L. Chiari}.}
  \bibinfo{year}{2012}\natexlab{}.
\newblock \showarticletitle{{Smartphone-Based Solutions for Fall Detection and
  Prevention: the FARSEEING Approach}}.
\newblock \bibinfo{journal}{\emph{Zeitschrift f{\"u}r Gerontologie und
  Geriatrie}} \bibinfo{volume}{45}, \bibinfo{number}{8} (\bibinfo{date}{Dec.}
  \bibinfo{year}{2012}), \bibinfo{pages}{722--727}.
\newblock
\urldef\tempurl%
\url{https://doi.org/10.1007/s00391-012-0404-5}
\showDOI{\tempurl}


\bibitem[Organization(2017)]%
        {WHO2017}
\bibfield{author}{\bibinfo{person}{World~Health Organization}.}
  \bibinfo{year}{2017}\natexlab{}.
\newblock \bibinfo{title}{Falls}.
\newblock
\newblock
\urldef\tempurl%
\url{https://www.who.int/en/news-room/fact-sheets/detail/falls}
\showURL{%
Retrieved April 26, 2017 from \tempurl}


\bibitem[Rashidi and Mihailidis(2013)]%
        {Rashidi2013}
\bibfield{author}{\bibinfo{person}{Parisa Rashidi} {and} \bibinfo{person}{Alex
  Mihailidis}.} \bibinfo{year}{2013}\natexlab{}.
\newblock \showarticletitle{{A Survey on Ambient-Assisted Living Tools for
  Older Adults}}.
\newblock \bibinfo{journal}{\emph{IEEE Journal of Biomedical and Health
  Informatics}} \bibinfo{volume}{17}, \bibinfo{number}{3} (\bibinfo{date}{Dec.}
  \bibinfo{year}{2013}), \bibinfo{pages}{579--590}.
\newblock
\urldef\tempurl%
\url{https://doi.org/10.1109/JBHI.2012.2234129}
\showDOI{\tempurl}


\bibitem[Romero-Tapiador et~al\mbox{.}(2023)]%
        {RomeroTapiador2023}
\bibfield{author}{\bibinfo{person}{Sergio Romero-Tapiador},
  \bibinfo{person}{Blanca Lacruz-Pleguezuelos}, \bibinfo{person}{Ruben
  Tolosana}, \bibinfo{person}{Gala Freixer}, \bibinfo{person}{Roberto Daza},
  \bibinfo{person}{Cristina~M Fernández-Díaz}, \bibinfo{person}{Elena
  Aguilar-Aguilar}, \bibinfo{person}{Jorge Fernández-Cabezas},
  \bibinfo{person}{Silvia Cruz~Gil}, \bibinfo{person}{Susana Molina-Arranz},
  \bibinfo{person}{Maria~Carmen Crespo}, \bibinfo{person}{Teresa Laguna-Lobo},
  \bibinfo{person}{Laura~Judith Marcos-Zambrano}, \bibinfo{person}{Ruben
  Vera-Rodriguez}, \bibinfo{person}{Julian Fierrez}, \bibinfo{person}{Ana
  Ramírez~de Molina}, \bibinfo{person}{Javier Ortega-Garcia},
  \bibinfo{person}{Isabel Espinosa-Salinas}, \bibinfo{person}{Aythami Morales},
  {and} \bibinfo{person}{Enrique Carrillo~de Santa~Pau}.}
  \bibinfo{year}{2023}\natexlab{}.
\newblock \showarticletitle{{AI4FoodDB: A Database for Personalized e-Health
  Nutrition and Lifestyle through Wearable Devices and Artificial
  Intelligence}}.
\newblock \bibinfo{journal}{\emph{Database: The Journal of Biological Databases
  and Curation}} (\bibinfo{year}{2023}).
\newblock


\bibitem[Schwickert et~al\mbox{.}(2013)]%
        {Schwickert2013}
\bibfield{author}{\bibinfo{person}{L. Schwickert}, \bibinfo{person}{C. Becker},
  \bibinfo{person}{U. Lindemann}, \bibinfo{person}{C. Mar{\'e}chal},
  \bibinfo{person}{A. Bourke}, \bibinfo{person}{L. Chiari},
  \bibinfo{person}{J.~L. Helbostad}, \bibinfo{person}{W. Zijlstra},
  \bibinfo{person}{K. Aminian}, \bibinfo{person}{C. Todd}, \bibinfo{person}{S.
  Bandinelli}, \bibinfo{person}{J. Klenk}, \bibinfo{person}{for~the
  FARSEEING~Consortium}, {and} \bibinfo{person}{the FARSEEING Meta Database
  Consensus~Group}.} \bibinfo{year}{2013}\natexlab{}.
\newblock \showarticletitle{{Fall Detection with Body-Worn Sensors}}.
\newblock \bibinfo{journal}{\emph{Zeitschrift f{\"u}r Gerontologie und
  Geriatrie}} \bibinfo{volume}{46}, \bibinfo{number}{8} (\bibinfo{date}{Dec.}
  \bibinfo{year}{2013}), \bibinfo{pages}{706--719}.
\newblock
\urldef\tempurl%
\url{https://doi.org/10.1007/s00391-013-0559-8}
\showDOI{\tempurl}


\bibitem[Sucerquia et~al\mbox{.}(2017)]%
        {Sucerquia2017}
\bibfield{author}{\bibinfo{person}{Angela Sucerquia},
  \bibinfo{person}{José~David López}, {and} \bibinfo{person}{Jesús~Francisco
  Vargas-Bonilla}.} \bibinfo{year}{2017}\natexlab{}.
\newblock \showarticletitle{{SisFall: A Fall and Movement Dataset}}.
\newblock \bibinfo{journal}{\emph{Sensors}} \bibinfo{volume}{17},
  \bibinfo{number}{1} (\bibinfo{date}{Jan.} \bibinfo{year}{2017}),
  \bibinfo{pages}{1--14}.
\newblock
\urldef\tempurl%
\url{https://doi.org/10.3390%2Fs17010198}
\showDOI{\tempurl}


\bibitem[Zhang et~al\mbox{.}(2015)]%
        {Zhang2015}
\bibfield{author}{\bibinfo{person}{Zhong Zhang}, \bibinfo{person}{Christopher
  Conly}, {and} \bibinfo{person}{Vassilis Athitsos}.}
  \bibinfo{year}{2015}\natexlab{}.
\newblock \showarticletitle{{A Survey on Vision-Based Fall Detection}}. In
  \bibinfo{booktitle}{\emph{Proceeding of the 8th ACM International Conference
  on Pervasive Technologies Related to Assistive Environments (PETRA '15)}}.
  {Association for Computing Machinery}, \bibinfo{address}{New York, NY, USA}.
\newblock


\bibitem[Zurbuchen et~al\mbox{.}(2021)]%
        {Zurbuchen2021}
\bibfield{author}{\bibinfo{person}{Nicolas Zurbuchen}, \bibinfo{person}{Adriana
  Wilde}, {and} \bibinfo{person}{Pascal Bruegger}.}
  \bibinfo{year}{2021}\natexlab{}.
\newblock \showarticletitle{{A Machine Learning Multi-Class Approach for Fall
  Detection Systems Based on Wearable Sensors with a Study on Sampling Rates
  Selection}}.
\newblock \bibinfo{journal}{\emph{Sensors}} \bibinfo{volume}{21},
  \bibinfo{number}{3} (\bibinfo{date}{Jan.} \bibinfo{year}{2021}).
\newblock
\urldef\tempurl%
\url{https://doi.org/10.3390/s21030938}
\showDOI{\tempurl}


\bibitem[Özdemir and Barshan(2014)]%
        {Ozdemir2014}
\bibfield{author}{\bibinfo{person}{Ahmet~Turan Özdemir} {and}
  \bibinfo{person}{Billur Barshan}.} \bibinfo{year}{2014}\natexlab{}.
\newblock \showarticletitle{{Detecting Falls with Wearable Sensors Using
  Machine Learning Techniques}}.
\newblock \bibinfo{journal}{\emph{Sensors}} \bibinfo{volume}{14},
  \bibinfo{number}{6} (\bibinfo{date}{June} \bibinfo{year}{2014}),
  \bibinfo{pages}{10691--10708}.
\newblock
\urldef\tempurl%
\url{https://doi.org/10.3390/s140610691}
\showDOI{\tempurl}


\end{thebibliography}

\end{document}